\def\year{2022}\relax
\documentclass[letterpaper]{article} 
\usepackage{aaai22}  
\usepackage{times}  
\usepackage{helvet}  
\usepackage{courier}  
\usepackage[hyphens]{url}  
\usepackage{graphicx} 
\usepackage{natbib}  
\usepackage{caption} 
\DeclareCaptionStyle{ruled}{labelfont=normalfont,labelsep=colon,strut=off} 
\usepackage{booktabs}
\usepackage[table,xcdraw]{xcolor}
\usepackage{comment}
\usepackage{xcolor,colortbl}
\usepackage[]{todonotes}
\usepackage{amsthm}
\theoremstyle{definition}
\newtheorem{definition}{Definition}
\newtheorem{thm}{Theorem}
\newtheorem{example}[thm]{Example}

\usepackage{newfloat}
\usepackage{algorithm}
\usepackage{algorithmic}

\title{Explainable Goal Recognition: A Framework Based on Weight of Evidence}
\author {
    Abeer Alshehri, \textsuperscript{\rm 1,3}
    Tim Miller, \textsuperscript{\rm 1}
    Mor Vered \textsuperscript{\rm 2}
}
\affiliations {
    \textsuperscript{\rm 1} University of Melbourne, Melbourne, Australia\\
    \textsuperscript{\rm 2} Monash University, Melbourne, Australia\\
    \textsuperscript{\rm 3} King Khalid University, Abha, Saudi Arabia\\
    aalshehri@student.unimelb.edu.au, 
    tmiller@unimelb.edu.au, 
    mor.vered@monash.edu
    }

\usepackage{bibentry}

\begin{document}

\maketitle

\begin{abstract}
We introduce and evaluate an eXplainable Goal Recognition (XGR) model that uses the Weight of Evidence (WoE) framework to explain goal recognition problems. Our model provides human-centered explanations that answer `why?' and `why not?' questions. We computationally evaluate the performance of our system over eight different domains. Using a human behavioral study to obtain the ground truth from human annotators, we further show that the XGR model can successfully generate human-like explanations. We then report on a study with 60 participants who observe agents playing Sokoban game and then receive explanations of the goal recognition output. We investigate participants’ understanding obtained by explanations through task prediction, explanation satisfaction, and trust. 

\end{abstract}

\noindent  
\noindent In recent years, a significant amount of research has been conducted on explainable AI (XAI) to increase the transparency of AI decision-making and improve the user's trust \cite{vered2020demand}. Although the main focus has been on Explainable Machine Learning, recently, there has been growing interest in Explainable Agency \cite{langley2017explainable,de2018explainable,hoffmann2019explainable,chakraborti2020emerging}; agents and robots capable of explaining their decisions to lay users. 
 
For a goal recognition (GR) problem, the task is to infer the most likely goal given an observed agent's behavior. For instance, when autonomous vehicles' anticipated goals are justified to end-users, it would assist to calibrate their trust in such systems \cite{shahrdar2018survey}. We are motivated by the necessity of generating human-like explanations for why a certain goal is most likely. 
Model-based GR approaches use domain models to generate plans for goals \cite{masters2021s}, and machine-learning  GR approaches rely on the existence of a  corpus of prior plans/observations from which to train \cite{pereira2019online}.
While there are many ways to achieve this objective, little to no attention has been given to explaining the output of these algorithms, once achieved.

There is a substantial body of literature in cognitive science that explores how humans explain others' behavior \cite{kashima1998category,malle2006mind,heider2013psychology}. People's view of the world is normally characterized by their beliefs, goals, and intentions. Reasoning over these mental states with causal relationships lies at the foundation of folk explanations of human behavior \cite{malle2000conceptual}. In light of the existing theory of behavior explanation \cite{malle2006mind}, \citeauthor{7343756b66944695b3c1358889845134} \shortcite{7343756b66944695b3c1358889845134} developed a conceptual framework grounded on a human study, with the aim to learn how people explain GR agent behavior; what concepts people use to generate explanations for answering `why' and `why not' questions. However, they have not built an explainability method. In this paper, we extend \cite{7343756b66944695b3c1358889845134}'s work and propose an \textit{eXplainable Goal Recognition model (XGR)} model that generates explanations consistent with the corresponding human explanation. 
 
We introduce a general XGR model based on the concept of Weight of Evidence (WoE) from information theory \cite{wod1985weight,melis2021human}. The model explains the output goal hypothesis of a GR algorithm by obtaining WoE values of observed behavior to determine to what extent an observation is responsible for one goal hypothesis in contrast to another. We define an explanation selection for `why goal $g$?' and `why not goal $g'$?' questions based on the concept of observational markers, the observation with the highest WoE, and counterfactual observational markers, the observation with the lowest WoE \cite{7343756b66944695b3c1358889845134}.

We computationally evaluate our approach on eight GR benchmark domains using a state-of-the-art GR model \cite{vered2018towards}. Results indicate that our model's computation time is a fraction of the original GR approach. We also conduct a follow-up human study in which participants were presented with the output of the GR model and asked to answer \textit{Why?} and \textit{Why not?} questions. We evaluate our model by comparing human-generated explanations to the output of the XGR model. Results show the efficiency of our model to generate human-like explanations. We conduct another human study using the proposed model for the GR agent that predicts a player's goal in the Sokoban game. Experiments were run for 60 participants, in which we evaluate the participants’ performance in task prediction, explanation satisfaction, and trust. Results show that our model has a better performance than the tested baseline. To the best of our knowledge, this is the first study to solve the problem of GR explainability more naturally and elegantly by adopting the concept of WoE.


%
%

%
%

\section{Related Work and Background}


\subsection{Explainable Agency}

A number of studies focus on generating explanations for action/activity recognition models and domains. Some recent examples include \cite{Meng2019-ly}, which uses LSTM based attention mechanism to identify the most relevant frames for video action recognition, and \cite{akula2022cx} which explains decisions made by a deep CNN over image recognition models. These approaches, as well as others, rely on machine learning to generate explanations rather than having an explicit model of the recognizing agent. 

Other approaches, such as \citeauthor{Albrecht2021-yh} \shortcite{Albrecht2021-yh} and \citeauthor{Brewitt2021-hq,brewitt2022verifiable} \shortcite{Brewitt2021-hq,brewitt2022verifiable} rely on the innate interpretability of the structure of their specific GR approach.  \citeauthor{Brewitt2021-hq} \shortcite{Brewitt2021-hq} utilize decision trees trained on vehicle trajectory data and \citeauthor{Albrecht2021-yh} \shortcite{Albrecht2021-yh} rely on inverse planning and Monte Carlo Tree Search. While not dependent on ML, these approaches also do not perform model-based explanations and are only adapted to one specific instance of a GR algorithm.

In the context of sequential decision-making, several explainable agency models have been proposed for Belief-Desire-Intention (BDI) agents \cite{cranefield2017no,winikoff2018bad}, reinforcement learning RL agents \cite{fukuchi2017autonomous,madumal2019explainable}, and planning agents \cite{chakraborti2017plan,chakraborti2018human,cashmore2019towards}.
These frameworks are mostly driven by goal-directed tasks over understanding the autonomous agents' decisions. These approaches, however, do not focus on GR agents.

Previous studies investigating the explainability of goal/intention recognition agent falls into the scope of maximizing the explicability of the agent behavior \cite{yolanda2015fast,sohrabi2016plan,vered2016online,Hu2021-ds,Hanna2021-rm}. This involves making that behavior more explicable to an observer by either aligning its behavior with the observer's expectations or making its inference formation interpretable.

\subsection{Planning}

Planning is a way to find a sequence of actions (i.e, a plan) that achieves a certain goal from an initial state. The concept of planning is key to understanding GR algorithms that utilize planners in the recognition process. Our \textit{eXplainable Goal Recognition model (XGR)} generates explanations for such GR models and also uses off-the-shelf planners to generate a counterfactual plan as part of that explanation.  We build upon the following planning problem definition as defined in \cite{pereira2017landmark}:

\begin{definition}
A planning task is represented by a triple $\langle \Xi, \mathcal{I},g \rangle$, in which $ \Xi = \langle \mathcal{F}, \mathcal{A} \rangle$ is a planning domain deﬁnition; $\mathcal{F}$ consists of a finite set of facts and $\mathcal{A}$ is a finite set of actions; $\mathcal{I}$ is the initial state, and $g$ is the goal state. A solution to a planning task is a plan $\pi$ that reaches a goal state $g$ from the initial state $\mathcal{I}$ by following a sequence of actions. 
Since actions have an associated cost, we assume that this cost is 1 for all actions. 
The objective is to find the optimal plan $\mathrm{\pi}_{}^{*}$ that minimizes the associated cost.
\end{definition}

\subsection{Goal Recognition (GR)}
Goal recognition (GR) is the inverse of the planning problem. It is the task of recognizing an agent's unobserved goal through a sequence of observations. There are many different approaches to solving the GR problem. Among the most common approaches are; library based GR algorithms that use dedicated plan recognition libraries that aim to represent all known ways to achieve known goals \cite{sukthankar2014plan};  
Model-based GR algorithms \cite{ramirez2010probabilistic,sohrabi2016plan,vered2016online} in which GR agents use their domain knowledge, represented through the use of planners, to generate plans that must be carried out for a goal to be achieved \cite{masters2021s}; and machine-learning GR approaches that rely on the existence of a large training corpus from which algorithms can learn about the constraints of the domain  \cite{min2014deep,pereira2019online,meneguzzi2021survey,fitzpatrick2021behaviour}. Among all of these approaches little, to no, attention has been given to explaining the reasons behind the predicted goals and/or goal probability distribution, which is the aim of this work. To begin we build on the following formal GR definition as defined by \cite{shvo2020active}.

\begin{definition}
A goal recognition problem is a tuple $\langle \Xi, \mathcal{I}, \mathcal{G}, \mathcal{O} \rangle$, in which $ \Xi = \langle \mathcal{F}, \mathcal{A} \rangle$ is a planning domain definition where $\mathcal{F}$ and $\mathcal{A}$ are sets of facts and actions, respectively; $\mathcal{I}$ is the initial state; $\mathcal{G} = \{ g_{1}, g_{2}, ..., g_{m}\}$ is the goals set, and $\mathcal{O} = \langle o_{1}, o_{2}, ..., o_{n}\rangle$ is a sequence of observations such that each $o_{i}$ is a pair $(\alpha_{i}, \phi_{i})$  composed of an observed action $\alpha_{i} \in \mathcal{A}$ and a fact set that represent the state $\phi_{i} \subseteq \mathcal{F}$. A solution to a GR problem is a probability distribution over $\mathcal{G}$ giving the corresponding likelihood of each goal, i.e. the posterior probability $P(g_{j} \mid \mathcal{O} )$ for each $g_{j} \in \mathcal{G}$. The most likely goal is the one whose generated plan “best satisfies” the observations.

\end{definition}


\subsubsection{The Mirroring GR Algorithm}

As part of our model's empirical evaluation, we will be explaining the output of the \textit{Mirroring} GR algorithm \cite{vered2017heuristic,kaminka2018plan}. The model has been inspired by humans' ability to do online GR which stems from the human brain's mirror neuron system for matching the observation and execution of actions \cite{rizzolatti2005mirror}. 
The approach belongs to the \textit{plan recognition as planning} GR approaches \cite{masters2021s} and utilizes a planner within the recognition process to compute alternative plans. In particular, the Mirroring algorithm calls a planner first to pre-compute optimal plans from $\mathcal{I}$ to every $g_{j} \in \mathcal{G}$ as well as to compute \textit{suffix} plans from the last observation $o_{i} \in \mathcal{O}$ to every $g_{j} \in \mathcal{G}$. These \textit{suffix} plans are concatenated with a \textit{prefix} plan (the observation sequence $\mathcal{O}$ at time step $t$) to generate new plan hypotheses. The algorithm then provides a likelihood distribution (posterior probabilities) over $\mathcal{G}$ by evaluating which of the generated plans, that incorporate the observations $\mathcal{O}$, best matches the optimal plan.


\begin{figure}[t]
\centering
\includegraphics[width=0.7\columnwidth]{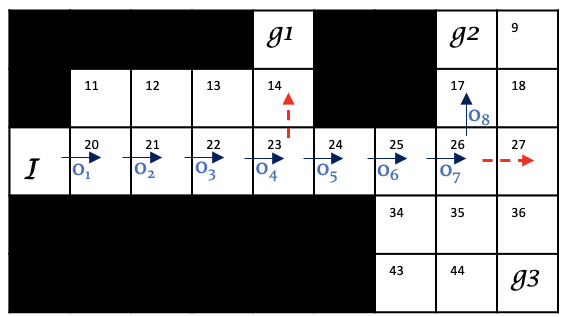} 
\caption{Navigational domain example. $\mathcal{I}$ is the agent's initial state. Given three possible goal locations ($g_1,g_2,g_3$), the predicted goal is $g_2$. Blue arrows represent the observation sequence, and red arrows represent counterfactual actions.}
\label{fig1}
\end{figure} 

\subsection{Running Example} 
\label{running_example}
To better explain the concepts previously introduced we present an MDP model of a navigational GR scenario, shown in Figure \ref{fig1}. In this scenario, an agent must navigate through the unblocked grid to reach one of three possible goal cells, $g_1, g_2$, or $g_3$. The state space $\mathcal{S}$ is defined by the cells, 45 states in total, and the initial state of the agent is at cell 19, labeled  $\mathcal{I}$. The action set $\mathcal{A}$ comprises moving one cell in each of the four directions (up, down, left, right) with equal cost, and transitions the agent between two connected cells. The GR problem in this example is composed of the following: the initial state, $\mathcal{I}$, set of goal hypotheses, $\mathcal{G} = \{g_1,g_2,g_3\}$, and a sequence of observations $\mathcal{O} = \langle o_1,..,o_8 \rangle$ at time step $t=8$ whose transition between them is represented as blue arrows. The problem here is deterministic, i.e., at a given state, each action leads to a determined state. As the goal state specification, $\mathcal{G}$ is for the agent to be at one of three possible goal cells. 
Over this example, the Mirroring GR algorithm would rank goal $g_2$ as most likely since the observation sequence confirms the optimal plan to achieve this goal. We will refer back to this example throughout the paper to show the performance of our approach on the output of the Mirroring GR algorithm.

\subsection{Weight of Evidence (WoE)}

The principle of rational action \cite{hempel2013rational} states that people explain goal hypotheses by determining to what extent each observed action is responsible for a goal hypothesis in contrast to others. Based on this, \citeauthor{bertossi2020asp} \shortcite{bertossi2020asp} defines a causal explanation to be the most responsible set of features for an outcome. Thus, we model our explanation model using the concept of Weight of Evidence.

Weight of Evidence (WoE) is a statistical concept used to describe variable effects in prediction models \cite{wod1985weight}. It has been defined in terms of log-odds (see supplementary material) to measure the strength of evidence $e$ in favor of a hypothesis $h$ and against an alternative hypothesis $h'$, conditioned on additional information $c$. Assuming uniform prior probabilities\footnote{See the supplementary material for the equations for when priors are not uniform.}, it is defined as: 

\begin{equation}
woe(h/h' : e \mid c) = \log \frac{P(h \mid e, c)}{P(h' \mid e, c)}  
\label{eq_woe}
\end{equation}

\citeauthor{melis2021human} \shortcite{melis2021human} propose a framework based on WoE for explaining machine learning classification problems and argue that this is a natural model that corresponds to the phenomena of how people explain to each other  \cite{miller2019explanation}. \citeauthor{melis2021human} \shortcite{melis2021human} found WoE naturally captures a contrastive statement, i.e. evidence for or against something. That would help answer questions like why goal $g$, why not goal $g'$, and what should have happened instead if the goal was $g'$. We adopt this concept and apply it to GR problems.

\section{eXplainable Goal Recognition (XGR)}

In this section, we present a simple and elegant explainability model for GR algorithms called \textit{eXplainable Goal Recognition} (XGR). Extending \citeauthor{melis2021human}'s \shortcite{melis2021human} WoE framework, our model answers `why' and `why not' questions, as they are the most demanded explanatory questions \cite{lim2009and}. 

Our explanation generation approach consists of two parts. The first part ranks each observation in an observed plan by its WoE score. The second part selects highly-ranked observations, obtaining a minimal complete explanation.

\subsection{Model overview}
Our explainable GR model accepts four inputs, which can be provided by any GR model: 
\begin{enumerate}
\item An observed sequence $\mathcal{O}$; 
\item A set of predicted goals, $G_p \subseteq \mathcal{G}$; 
\item 
A set of counterfactual (not predicted) goals, $G_c \subset \mathcal{G}$, whereby $G_p \cap G_c = \emptyset$;  and
\item  Posterior probabilities $P(g \mid \mathcal{O})$ for each $g \in \mathcal{G}$. 
\end{enumerate}
 The model answers two questions: `Why goal $g$?', where $g$ is a predicted goal hypothesis; and `Why not goal $g'$?', where $g'$ is a counterfactual goal hypothesis.

\subsection{Explanation Generation}

We define an explanation as a pair containing: (1) an explanandum, the event to be explained; and (2) an explanan, a list of causes given as the explanation \cite{miller2019explanation}. The explanandum is assumed to be of the form \textit{why/why not $g$?}, where $g$ is a goal. We extend the WoE framework \citep{melis2021human} to GR problems.

Referring to Equation~\ref{eq_woe}, we substitute the hypotheses $h$ and $h'$ with a predicted goal and counterfactual goal hypotheses, $g$ and $g'$, the evidence $e$ with the observation $o_i \in \mathcal{O}$, the additional information $c$ with the observed sequence $\mathcal{O}$ up to the observation $o_i$, and the posterior probabilities as $P(g \mid \mathcal{O}_i)$ and $P(g' \mid \mathcal{O}_i)$, in which $g \in G_p$ and $g' \in G_c$. We define a complete explanan as follows.

\begin{definition} A \emph{complete explanan} for a goal $g$ is a list of pairs $(woe(g/g' : o_i \mid \mathcal{O}), o_{i})$, in which the conditional $woe(g/g' : o_i \mid \mathcal{O})$ for each paired hypothesis $g$ and $g'$ is computed for each added observation $o_i$ to the observed sequence $\mathcal{O}$ each time step. The WoE is computed using:

\begin{equation}
woe(g/g' : o_i \mid \mathcal{O}) = \log \frac{P(g \mid \mathcal{O}_i)}{P(g' \mid \mathcal{O}_i)}  
\end{equation}

\end{definition}

Informally, this defines a complete explanan for a goal $g$ as the complete list of computed WoE scores for each observation. An algorithm for extracting this is shown in Algorithm~\ref{alg:algorithm}.

\begin{algorithm}
\caption{Explanation Generation Algorithm}
\label{alg:algorithm}

\textbf{Input}: $\mathcal{O}$, $G_p$, $G_c$, and $Posterior$ $probability$ over $\mathcal{G}$\\
\textbf{Output}: Explanation list $\Omega$ of all $G_p$ paired with $G_c$
\begin{algorithmic}[1] 
\FOR{$o_i \in \mathcal{O}$}
\STATE $\Omega \leftarrow \lbrack \rbrack$
\FOR{\textbf{each} \texttt{$g \in G_p$}}
 \FOR{\textbf{each} \texttt{$g' \in G_c$}}
\STATE \texttt{$\omega_{i} \leftarrow woe(g / g' : o_i \mid\mathcal{O})$} \\
\STATE \texttt{$\Omega \leftarrow \lbrack (g, g') = \langle \omega_{i}, o_{i} \rangle \rbrack$} 
  \ENDFOR
 \ENDFOR
\ENDFOR
\STATE \textbf{return} $\Omega$
\end{algorithmic}
\end{algorithm}


\begin{example}
In the navigational GR scenario presented in Figure~\ref{fig1}, a complete explanan has been generated for the observation sequence, $\mathcal{O} = \langle o_1,..,o_8 \rangle$. For the first four observations, $o_{1}$ to $o_{4} \in \mathcal{O}$, the WoE would be the same for all goal hypotheses. This is because the Mirroring GR algorithm predicts them as equally likely since these observations are part of the optimal plan to achieve all three goals. 
 
 However, this will not be the case for the rest of the observation sequence. For observations $o_{5}$ to $o_{7} \in \mathcal{O}$, the Mirroring GR algorithm output would be goals $g2$ and $g3$ (both equally likely), and the counterfactual goal would be $g1$. After observation, $o_{8}$, the predicted goal would be $g_2$, and the counterfactual goals would be $g1$ and $g3$. Table~\ref{woetab} presents the generated complete explanan with each new observation.
\end{example}
%

\begin{table}[!h]
\centering
\small
\begin{tabular}{@{}rrrr@{}}
\toprule
\multicolumn{1}{c}{$o_i$} & \multicolumn{1}{c}{$(g, g')$} & \multicolumn{1}{c}{$\omega_i$} & \multicolumn{1}{c}{$\Omega$} \\ \midrule
$o_5$ & $(g_2, g_1)$ & 0.28 & $\langle0.28, o_5\rangle$ \\
 & $(g_3, g_1)$ & 0.28 & $\langle0.28, o_5\rangle$ \\ \midrule
$o_6$ & $(g_2, g_1)$ & 0.51 & $\langle0.51, o_6\rangle$ \\
 & $(g_3, g_1)$ & 0.51 & $\langle0.51, o_6\rangle$ \\ \midrule
$o_7$ & $(g_2, g_1)$ & 0.69 & $\langle0.69, o_7\rangle$ \\
 & $(g_3, g_1)$ & 0.69 & $\langle0.69, o_7\rangle$ \\ \midrule
$o_8$ & $(g_2, g_1)$ & 0.85 & $\langle0.85, o_8\rangle$ \\
 & $(g_2, g_3)$ & 0.18 & $\langle0.18, o_8\rangle$ \\ \bottomrule
\end{tabular}
\caption{Complete explanan WoE for predicted and counterfactual goals after observations $o_4,o_5$ and $o_6$ in the navigational GR example depicted in Figure ~\ref{fig1}}
\label{woetab}
\end{table}
\subsection{Explanation Selection}
The task of explaining a GR algorithm's output in terms of the complete explanan would be tedious or even impossible, particularly in a domain where an MDP model contains hundreds of thousands of states and actions. Indeed, `good' explanations should be selective by focusing on one or two possible causes instead of all possible causes for a decision or recommendation \cite{miller2019explanation}. In the context of GR, people explain in terms of the most important observation to achieve certain goals in contrast with other alternatives \cite{7343756b66944695b3c1358889845134}. To this end, we focus on the explanation selection of answering `Why $g$?' and `Why not $g'$?' questions.

\subsection{`Why' questions}
Answering \textit{why  goal $g$?} questions, as in, \textit{Why is goal $g$ predicted as the most likely goal candidate?}  relies on identifying the most important observation(s) that support the achievement of that goal. Following \citeauthor{7343756b66944695b3c1358889845134} \shortcite{7343756b66944695b3c1358889845134}, we call such observations \emph{observational markers}.

\begin{definition} Given a complete explanan of $g$, the \emph{observational markers} (\textit{OMs}) are the observation(s) that have the highest WoE value:
\begin{equation}
OM = \arg\max_{o_{i} \in \mathcal{O}} [(g, g') = \langle \omega_{i}, o_{i} \rangle]
\end{equation}
\end{definition}

There may be multiple such observations, in which case we select them all.

\begin{example}
Let us go back to the navigational GR scenario and answer the question \textit{Why $g_2$?}. From the \textit{complete explanan} of $g_2$, shown in Table ~\ref{woetab}:
\[
\begin{array}{lll}
 (g_2, g_1) = & [\langle0.28, o_5\rangle, \langle0.51, o_6\rangle, \langle0.69, o_7\rangle,\langle0.85, o_8\rangle]  \\ 
 (g_2, g_3) = & [\langle0.18, o_8\rangle]  \\ 
\end{array}
\]
    After ranking them from highest to lowest, we obtain $\langle0.85, o_8\rangle$ that has the highest value. This indicates that this observation is the $OM$, as in the observation that best explains the predicted goal hypothesis $G_p = \{g_2\}$ instead of the counterfactual goal hypotheses, $G_c = \{g_1, g_3\}$. Therefore the explanation would be \textit{Because the agent has moved up from cell 26 to cell 17}.
\end{example}



\subsection{`Why Not' questions}

The question of \textit{why not $g'$} relies on identifying the most important observation(s) to $g'$, which \citeauthor{7343756b66944695b3c1358889845134} \shortcite{7343756b66944695b3c1358889845134} call a \emph{counterfactual observational markers}.

\begin{definition} Given a complete explanan of $g'$, the \emph{counterfactual observational markers} (\textit{counterfactual OMs}) are the observation(s) that have the lowest WoE value:
\begin{equation}
counterfactualOM = \arg\min_{o_{i} \in \mathcal{O}} [(g, g') = \langle \omega_{i}, o_{i} \rangle]
\end{equation}
\end{definition}

\begin{example}
Let us go back to the navigational GR scenario and answer the question \textit{Why not $g_1$ and $g_3$?}. From the \textit{complete explanan} of $g_1$ and $g_3$, shown in Table ~\ref{woetab}:
\[
\begin{array}{lll}
 (g_2, g_1) = & [\langle0.28, o_5\rangle, \langle0.51, o_6\rangle, \langle0.69, o_7\rangle,\langle0.85, o_8\rangle]  \\ 
 (g_2, g_3) = & [\langle0.18, o_8\rangle]  \\ 
\end{array}
\]
        After ranking them from lowest to highest, we obtain $\langle0.28, o_5\rangle$ that has the lowest value of $g_1$ and $\langle0.18, o_8\rangle$ that has the lowest value for $g_3$. This indicates that these observations are the $counterfactualOM$, as in the observations that best explain the counterfactual goal hypotheses, $G_c = \{g_1, g_3\}$. Therefore the explanation would be \textit{Because the agent has moved right from cell 23 to cell 24 away from $g_1$, and it has moved up from cell 26 to cell 17 away from $g_3$}.

\end{example}

\subsubsection{Counterfactual Action}

Pointing to the lowest WoE action is not such a useful way to understand why a counterfactual goal is not predicted. \citeauthor{7343756b66944695b3c1358889845134} \shortcite{7343756b66944695b3c1358889845134} show that part of being able to answer `why not' questions is the ability to reason about the counterfactual actions that should have happened instead of \textit{counterfactual OM} for $g'$ to be the predicted goal \cite{7343756b66944695b3c1358889845134}. It is called \emph{counterfactual plan}.

Building on this idea, we obtain the counterfactual plan that should have happened instead of the observed one by planning the agent route to $g'$, and simply taking the first action.  We approach this problem by generating a plan for $g1$ from the state that precedes obtaining the \textit{counterfactual OM}, the state from which the lowest WoE is measured. We define the counterfactual action as follows.

\begin{definition} Given a \textit{counterfacual OM} $o_i$ at state $s_{t-1}$ for counterfactual goal $g'$, a \emph{counterfactual action explanation} is the action $a'_{t}$, which is the first action from the plan $\pi = (a'_{t}, a'_{t+1}, . . , g')$ generated by solving the planning problem $\langle \textit{M}, s_{t-1}, g'\rangle$, where $\textit{M}$ is the planning domain.
\end{definition}
 
\begin{example}
Consider again the example from Figure~\ref{fig1}. The counterfactual action $a'_t$ would be the \textit{move up} action from cell 23 to 14 for $g_1$, and the \textit{move right} action from cell 26 to 27 for $g_3$ (as presented by the red arrows). Verbally, the complete explanation to `Why not goal $g_1$ and$g_3$?' would be \emph{because the agent moved right from cell 23 to cell 24, it would have moved up from cell 23 to 14 if the goal was $g_1$. And it has moved up from cell 26 to cell 17, it would have moved right from cell 26 to cell 27 if the goal was $g_3$}.
\end{example}



\section{Computational Evaluation}


We evaluate the computational cost of the XGR model over eight online GR benchmark domains \cite{vered2018towards}. 
The benchmark domains vary in the levels of complexity and size including the different number of observations and goal hypotheses. We measure the overall time taken to run the XGR model. As the explanation model uses an off-the-shelf planner for counterfactual planning, we also  separate the cost of the planner and the explanation generation and show what effect it has on overall model performance.

Table~\ref{tab1} presents the run time performance of the XGR model over the benchmark domains. The run times vary greatly depending on the complexity of the domain, ranging from an average of 0.14 seconds over the 15 problems in the relatively simple, Kitchen domain, to 221.77 seconds over the 16 problems in the complex, Zeno-Travel domain (column 1).  Regardless of the run time, adding our explainability model to the GR approach is typically not expensive, adding an increase of between 0.2\%-45\% (column 3).  However, most of this increase can be attributed to calling the planner to generate counterfactual explanations (column 4). We can see that between 70\%-99\%  of the XGR model is spent on planning. The varying percentage increases between domains like Zeno-Travel and Kitchen emphasize the relation between the domain complexity and planning time, the higher the domain complexity, the higher influence the planner has. This highlights the impact that planner choice can have on our model performance. As the XGR approach is independent of the underlying GR model, this also shows that the proposed model would scale well with more efficient planners; e.g.\ domain-specific planners.

\begin{table}[h]
\resizebox{\columnwidth}{!}{
\begin{tabular}{@{}lrrrr@{}}
\toprule
\textit{\begin{tabular}[c]{@{}r@{}}Domain   \\      (\# problems)\end{tabular}} &
  \begin{tabular}[c]{@{}r@{}}Mirroring with \\      XGR (sec)\end{tabular} &
  \begin{tabular}[c]{@{}r@{}}XGR \\only (sec) \end{tabular} &
  \begin{tabular}[c]{@{}r@{}}Time \\ Increase (\%)\end{tabular} &
  \begin{tabular}[c]{@{}r@{}}Counterfactual \\    Planning (\%)\end{tabular} \\ \midrule
\begin{tabular}[c]{@{}l@{}}Campus\\       (15)\end{tabular}      & 0.21 (0.08)   & 0.019 (0.017)  & 10.15 & 87.11 \\
\begin{tabular}[c]{@{}l@{}}Ferry \\      (24)\end{tabular}       & 71.22 (36.16)  & 6.276 (8.070) & 09.66 & 99.69 \\
\begin{tabular}[c]{@{}l@{}}Intrusion \\      (45)\end{tabular}   & 0.69 (0.36)  & 0.215 (0.087) & 44.61 & 70.18 \\
\begin{tabular}[c]{@{}l@{}}Kitchen \\      (15)\end{tabular}     & 0.14 (0.07)   & 0.014 (0.002) & 11.12 & 73.61 \\
\begin{tabular}[c]{@{}l@{}}Rovers \\      (20)\end{tabular}      & 135.23 (73.11)  & 3.710 (7.271) & 02.82 & 99.64 \\
\begin{tabular}[c]{@{}l@{}}Satellite \\      (27)\end{tabular}   & 16.76 (10.05)  & 1.794 (10.052) & 11.98 & 99.27 \\
\begin{tabular}[c]{@{}l@{}}Miconic \\      (20)\end{tabular}     & 109.12 (22.61)  & 1.636 (2.861) & 01.52 & 98.72 \\
\begin{tabular}[c]{@{}l@{}}Zeno-Travel \\      (16)\end{tabular} & 221.77 (68.85) & 8.856 (11.721) & 04.15  & 99.65 \\ \bottomrule
\end{tabular}}
\caption{Performance results of the XGR model, generating explanations for the \textit{Mirroring} GR over eight benchmarks. Column 1 shows the mean and standard deviation run time of XGR model with the mirroring GR. Column 2 shows the average and standard deviation run time of XGR model only, without the GR algorithm. 
Column 3 shows the increase in run time (as a percentage) of adding the XGR
to the GR. Column 4 shows how much time (as a percentage of column 3) of the XGR model was spent in counterfactual planning.
}
\label{tab1}

\end{table}



\section{Empirical Evaluation: Human Study}

\subsection{Human Study 1: Ground truth }
As there is no other explainable GR approach we do not have a baseline against which to compare the explanations of our approach. We, therefore, conducted a human subject study in which participants were presented with the output of the Mirroring GR algorithm \cite{vered2018towards} and required to answer questions about its recognition process. By comparing human-generated explanations to the output of our XGR model, we aim to evaluate whether the model output is grounded on human-like explanations. 

\subsubsection{Methodology} We presented participants with the Mirroring GR algorithm output over a range of problems in the Sokoban game domain, a classic warehouse puzzle game. To evaluate our hypothesis, we used the method of annotator agreement and ground truth whereby human annotation of representative features provides the ground truth for quantitative evaluation of explanation quality \cite{mohseni2018human}. We evaluated our XGR model by comparing it against this ground truth. Ethics approval was obtained from our institute before the study was performed.

\subsubsection{Experiment Design}  
\begin{figure}[h]
\centering
\includegraphics[width=0.9\columnwidth]{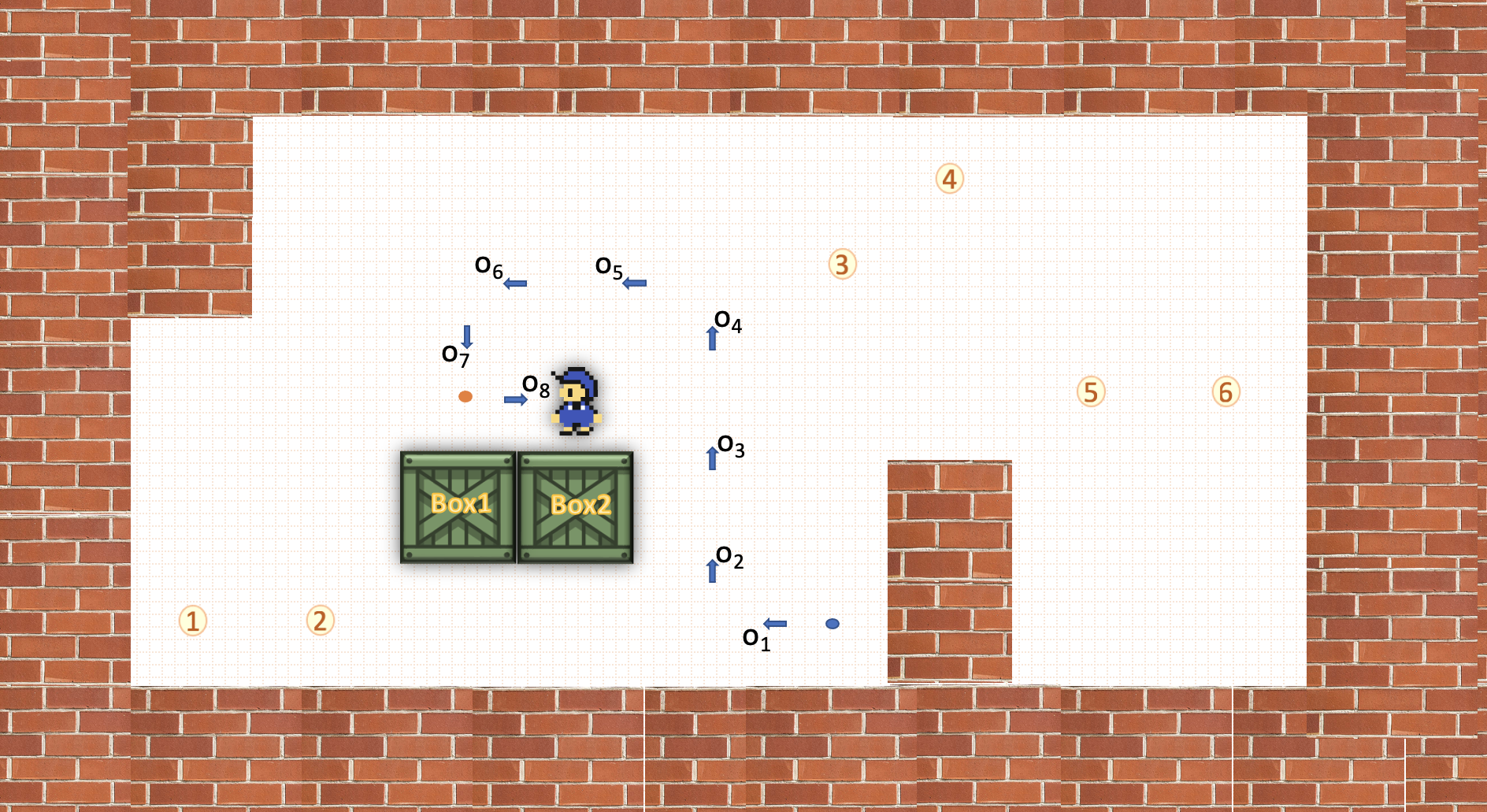} 
\caption{Sokoban game, scenario 5 (game version 3). The blue spot marks the initial state of the agent, and the orange spot marks the initial state of the box1 before pushing it down. There are 3 possible goals $((g_1,g_2),(g_3,g_4),(g_5,g_6))$. Blue arrows represent observations (8 observations). The predicted goal is $(g_1,g_2)$.}
\label{fig2}
\end{figure}

We built a modified version of the Sokoban game (the interface is shown in Figure~\ref{fig2}). Sokoban is a well-known, puzzle game in which a player moves boxes around a warehouse and delivers them to target storage locations. For our purpose, we modified the game rules by enabling the player to push more than one box simultaneously. This made predicting the target goal non-trivial.

To obtain a human explanation, i.e. `ground truth', over the  Sokoban GR task, we asked three annotators (one male, two female) recruited from the graduate student cohort at our university to annotate 15 scenarios, along with their preferred explanations. Participants were aged between 29 to 40 ($Mean = 33$). No prior knowledge was required. Each experiment ran for approximately 60 minutes. The annotation was conducted over four stages:
\begin{enumerate}
  \item The game instructions were introduced to the annotator with a training scenario to help them understand the task. 

  \item The annotator watched a partial scenario (video clip) in which a Sokoban player tried to achieve a goal. The goals were either delivering/pushing a box to a single destination cell or delivering/pushing two boxes to two different destination cells.

  \item After watching the observations, annotators were given the set of predicted goals, and counterfactual goals, that were predicted by the Mirroring GR algorithm. 
  
  \item The annotators were asked to annotate the most important observation, or optionally the two most important observations, from the observation sequence that answered the two questions: `Why goal $g$?' and `Why not goal $g'$?', where $g$ was the predicted goal and $g'$ was the counterfactual goal. Participants were also required to annotate a counterfactual action for `Why not goal $g'$?'. This was obtained by asking the participant to propose a \textit{non-observed} action that they believed would signify a move to the alternative goal $g'$.
  
\end{enumerate}

The data was collected over three game versions: one version required the delivery of a single box to one destination (game version 1), or two sequential destinations with interleaved plans to achieve them (game versions 2 and 3). The difference between game versions 2 and 3 pertained to the agent's ability to push multiple boxes in game 3, whereas in game 2 the agent could only push one box at a time. Each version was comprised of five different scenarios varying by complexity (15 scenarios in total). Each scenario depicted a different GR problem in which there were several competing goal hypotheses. The final destinations were not disclosed to the participants.

We combined the three annotations into a single ground truth using a majority vote. In the case of disagreements between annotations, they are often resolved by adjudicating by a fourth annotator. In our experiments, there were no disagreement instances.

We ran our model over the online GR mirroring implementation model for each of the 15 scenarios. We obtained the explanan list of XGR model to answer the \textit{Why goal $g$?} question by selecting the observations with the highest WoE ($OM$), and the \textit{Why not goal $g'$?} question by selecting the observations with the lowest WoE (\textit{counterfactual $OM$}). We then compared the output explanations of our model to the ground truth obtained by human annotation. To do this, each observation in the ground truth was given its equivalent rank in the model-generated ranked explanan list.

\begin{example}

Considering the example from Figure \ref{fig2}, we obtain the complete explanan for both questions from our model. The annotated observations from the `ground truth' is $o_{7}$ and $o_{2}$ that explain `Why $(g1,g2)$?', and $o_{2}$ that explains `Why not $((g3,g4)(g5,g6))$?'. We then assigned rank values to them based on the explanan list of each question, that is for why, rank values start from the highest, thus the annotated observation rank is $o_{2}$ = 7, and $o_{7}$ = 2 ($o_7$ rank value is 2 as it has the second highest WoE after $o_8$ that has rank value 1). For why not, rank values start from the lowest, thus the annotated observation rank is $o_{2}$ = 1 ($o_{2}$ is the lowest with rank value 1 as it has the lowest woe).

\end{example}

We then calculated the mean absolute error (MAE) to analyze how close the obtained explanation of the XGR model was to the ground truth. The MAE was calculated as the differences between each ground truth value ($a_{groundTruth}$) and XGR value ($a_{XGR}$) for the instance, averaged on the length of the observation sequence ($n$).


\begin{equation}
MAE = \frac{1}{n}\sum_{i=1}^{n}\mid a_{groundTruth} - a_{XGR}\mid
\end{equation}

The MAE is the average of the errors, hence the larger the number, the larger the error. An error of 0 indicates full agreement between the models, and an error of 1 means that the top-ranked observation was ranked last by the XGR model.

For evaluating the selection of a counterfactual action, as there is only one counterfactual action per plan, this is a binary agreement between ground truth and the XGR model. For each domain, we calculated the percentage of agreements using \cite{araujo1985calculating}:

\begin{equation}
CF (\%) = \frac{agreements}{agreements + disagreements}\times 100\%
\end{equation}

\subsubsection{Results}  

The results of the comparison are presented in Table ~\ref{tab:humanStudyResults}. 
Each row  shows the MAE value calculated for each game scenario. The `Why' and `Why not' columns represent the MAE for our model compared to the human ground truth, and the CF(\%) column represents the percentage of counterfactual action explanations that agreed with the human ground truth.
We can see that for the majority of instances, the XGR model agreed completely with the ground truth obtained through human annotation. When answering \textit{Why $g$?} questions the model agreed with the ground truth over 11 of the 15 scenarios (73.3\%) and when answering \textit{Why not $g'$?} questions the model agreed with the ground truth over 14 of the 15 scenarios (93.3\%). 

The CF column represents the percentage of counterfactual explanations that agreed with the human ground truth, so in this instance, higher values are better, with 100\% indicating full agreement. The full agreement rate was obtained in 11 of the 15 scenarios (73.3\%). 

 

\begin{table}[h]
\resizebox{\columnwidth}{!}{
\tiny
\centering
\begin{tabular}{lcccr} 
\toprule
\textbf{Game version} & \textbf{Scenario} & ~\textbf{Why}~~ & ~\textbf{Why Not}~ & \textbf{CF} (\%)  \\ 
\midrule
1                                                           & S1        & 0.00            & 0.00              & 100                                                                  \\
                                                                & S2        & 0.00            & 0.00              & 100                                                                  \\
                                                                & S3        &\textbf{ 0.37 }           & 0.00              & 100                                                                  \\
                                                                & S5        &\textbf{ 0.25 }           & 0.00              & 100                                                                  \\
                                                                & S5        & 0.00            & 0.00              & 100                                                                  \\[1mm]
2                                                         & S1        & 0.00            & 0.00              & 66.6                                                                \\
                                                                & S2        & 0.00            & \textbf{0.12}              & 33.3                                                                \\
                                                                & S3        & 0.00            & 0.00              & 100                                                                  \\
                                                                & S4        & 0.00            & 0.00              & 100                                                                  \\
                                                                & S5        & 0.00            & 0.00              & 33.3                                                                \\[1mm]
3                                                         & S1        & \textbf{0.50}            & 0.00              & 100                                                                  \\
                                                                & S2        & 0.00            & 0.00              & 50                                                                   \\
                                                                & S3        & 0.00            & 0.00              & 100                                                                  \\
                                                                & S4        & 0.00            & 0.00              & 100                                                                  \\
                                                                & S5        & \textbf{0.44}            & 0.00              & 100                                                                  \\ 
\midrule
Mean                                                            &           & 0.10            & 0.008             & 89.40                                                               \\
SD                                                              &           & 0.65            & 0.031             & 00.25                                                               \\
\bottomrule
\end{tabular}}
\caption{The \textit{Why} and \textit{Why not} columns represent the mean absolute error (MAE) for our model compared to the human ground truth. The CF columns represent the percentage of counterfactual action explanations that agreed with the human ground truth.}

\label{tab:humanStudyResults}
\end{table}

Investigating scenario 5 in game version 3, which has a relatively high MAE for the \textit{Why goal $g$?} question. The scenario can be seen in Figure~\ref{fig2} and involves the agent delivering 2 boxes to 2 different locations, while also being able to push 2 boxes at the same time. The blue arrows represent the observation sequence, whereby the agent started at the blue circle and moved along the arrows to its current location.  

In this scenario the most likely goal candidate, as predicted by the GR was delivering Box1 to $g_1$ and delivering Box2 to $g_2$, i.e. $G_p = \{(g_1,g_2)\}$. The counterfactual goal candidates were delivering the boxes to either $g_3$ and $g_4$  or $g_5$ and $g_6$, $G_c = \{(g_3,g_4), (g_5,g_6)\}$. It is important to note that in order to push both boxes simultaneously to goal $(g_1,g_2)$, the agent would need to stand on the right of the boxes, but to push both boxes simultaneously to either $(g_3,g_4)$ or $(g_5,g_6)$, it needs to position itself on the left.

The XGR model's explanation to the \textit{Why goal $g$?} question is the observation $o_8$. This can be seen as the agent aiming to position itself on the right of both boxes, confirming the supposition that the goal is $(g_1,g_2)$. Fully considering the agent's ability to push multiple boxes, this observation constitutes the $OM$, the observation with the highest WoE. 

On the other hand, the annotators established the ground truth explanation by choosing the second observation, $o_{2}$ for both the \textit{Why goal $g$?} and \textit{Why not $g'$?} questions. According to our model, this observation is the one with the lowest WoE, actually making it the \textit{counterfactual $OM$} and the answer to the question \textit{Why not $g'$?}. This is because this observation moves away from both goals $(g_3,g_4)$ and  $(g_5,g_6)$. Participants choose to use the same answer for both \textit{Why goal $g$?} and \textit{Why not $g'$?} questions can also be found in the other instances of discrepancies between the output of our model and the ground truth. The difference in explanations can be attributed to some confusion and/or preference between \textit{why?} and \textit{why not?} questions on the side of participants. Thus, we had a follow-up experiment where we presented the participants with the scenarios they had confusion with (Table \ref{tab:humanStudyResults}, scenarios of bold values). For each scenario, we provided them with two explanation systems to answer the two questions where the first system explanation is given by our model, and the second system explanation is given by the ground truth. Then we asked them which system provide a better explanation. For the three participants, the chosen system was the first one (our model).



%

\subsection{Human Study 2: Explainability}
We conduct  a second human subject experiment to evaluate the explainability of our model. We consider two hypotheses for our evaluation; 1) our model (XGR) leads to a better understanding of a GR agent; and 2) a better understanding of an agent fosters user trust in the agent.

\subsubsection{Methodology} We used the Sokoban game as the domain of the GR agent. We presented participants with the Mirroring GR algorithm output over six problems in Sokoban game domain. To evaluate hypothesis 1, we used the task prediction method \cite{hoffman2018metrics}. Task prediction is a proxy measure for user understanding. Participants are instructed to predict the goals of the agent. We used the \textit{explanation satisfaction scale} by \citeauthor{hoffman2018metrics} \citeyearpar{hoffman2018metrics} to measure the explanation's subjective quality. To evaluate hypothesis 2, we used the \textit{trust scale} by \citeauthor{hoffman2018metrics} \citeyearpar{hoffman2018metrics}.

\subsubsection{Experiment Design}
We presented six partial scenarios (video clips) of the Sokoban player trying to achieve the goal of delivering boxes to certain locations. The independent variable in this experiment is the explanation type: (1) our explanation model (XGR); and a baseline of no explanation. There is no baseline of another explanation method since there is no other method as far as we know.

The experiment has four phases. The first phase involves a collection of demographic information and training the participants. The participant is trained to understand the player task and how the GR system works using two video clips. In the second phase, a video clip (10 sec) is played and the GR system output is shown. Participants were asked to answer ``what is the agent's predicted goal(s)?''.  For the baseline condition, they answered it without receiving any  explanations. In the XGR condition, our model explanations  for `why’ and `why not’ questions were presented. Participants completed six scenarios each. The explanations are pre-generated from our implemented algorithm and displayed on an annotated picture of the video clip's last frame. In the third phase,  participants were asked to complete the trust scale. The second condition has a fourth phase of completing the explanation satisfaction scale. 

We conducted the experiments on Amazon MTurk with 60 participants, allocated randomly and evenly to each condition. Each experiment ran approximately 20 minutes, and we compensated each participant with \$4USD. A bonus compensation of \$0.20USD was given for each correct prediction, for a total of \$1.20USD. Participants were aged between 31 to 60 ($Mean = 41.4$), 23 were male while 37 were female; none selected non-binary or self-specified. To ensure data quality, we recruited only ‘master class’ workers with 98\% or more approval rates. We excluded 22 participants' answers as they completed the task in under two minutes.

\begin{figure}[ht]
\begin{minipage}[b]{0.45\linewidth}
\centering
\includegraphics[width=\textwidth]{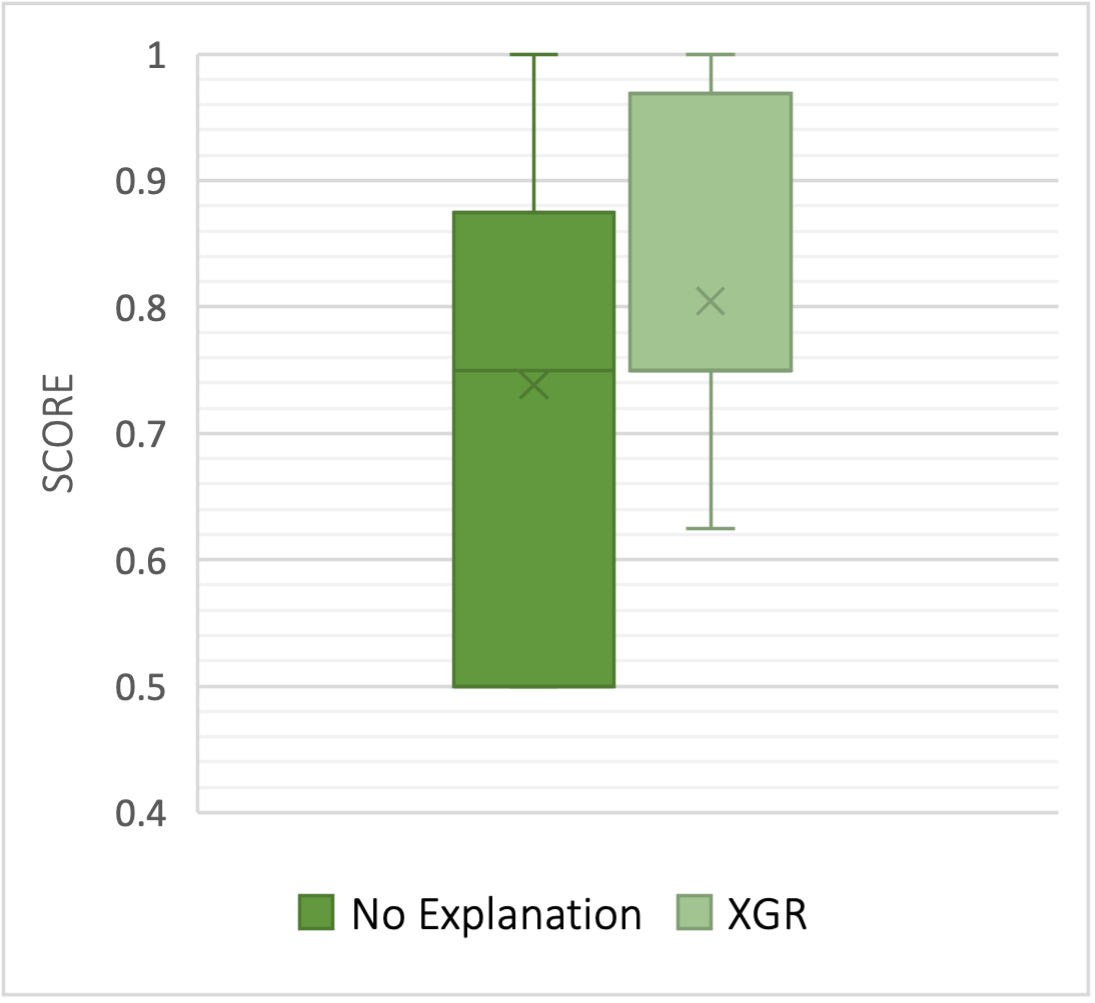}
\caption{Task prediction scores of the two models considering the correct response with means represented as markers (higher is better).}
\label{fig3}
\end{minipage}
\hspace{0.5cm}
\begin{minipage}[b]{0.45\linewidth}
\centering
\includegraphics[width=\textwidth]{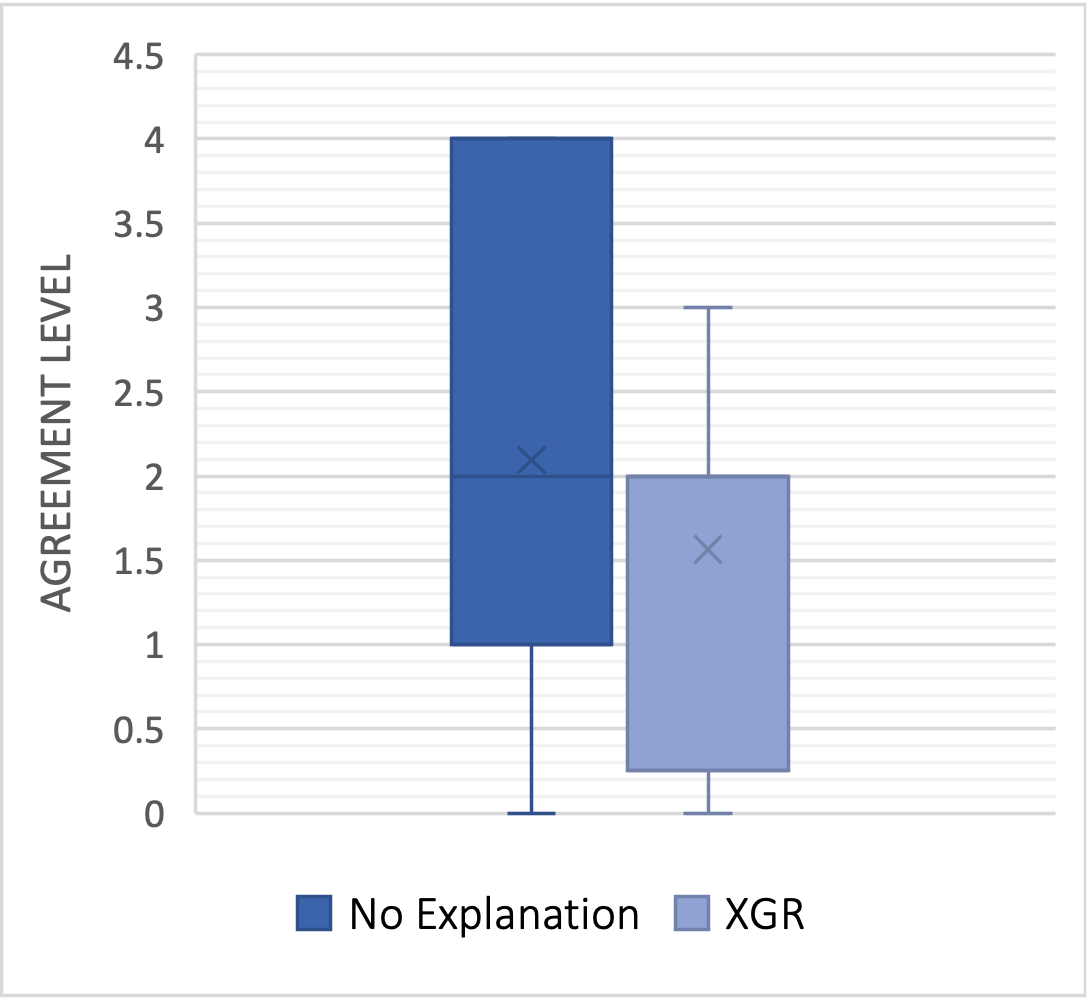}
\caption{Behavioural trust for the two models (lower is better)}
\label{fig4}
\end{minipage}
\end{figure}

\subsubsection{Results}

We first show our results on the first hypothesis, corresponding null and alternative hypothesis are, $H_0: S_{XGR} = S_{NoExplanation}$; and $H_1: S_{XGR} \geq S_{NoExplanation}$ where $S$ denotes the participants' prediction scores.

Figure (\ref{fig3}) shows the task score variance with the two models. A Welch two-sample t-test resulted in $p$-value of \textbf{0.05}, thus we reject $H_0$ and accept the alternative hypothesis $H_1$. These results demonstrate that the XGR model leads to a significantly better understanding of the agent's behavior than the baseline model. 

\begin{table}[h]
\centering
\small
\begin{tabular}{@{}llll@{}}
\toprule
Understanding & Satisfying  & Sufficient  & Complete    \\ \midrule
88.93 (10.8)  & 86.09 (12.2) & 87.93 (13.1) & 86.15 (19.6) \\ \bottomrule
\end{tabular}
\caption{Mean and standard deviation of explanation quality metrics for XGR model}
\label{tab:eXpmetrics}
\end{table}

Table~\ref{tab:eXpmetrics} shows the average and standard deviation of the explanation satisfaction metrics on a Likert scale percentage (a higher value indicates stronger agreement). These results indicate a satisfactory level across the four metrics.

\begin{figure}[h]
\centering
\includegraphics[scale=0.45]{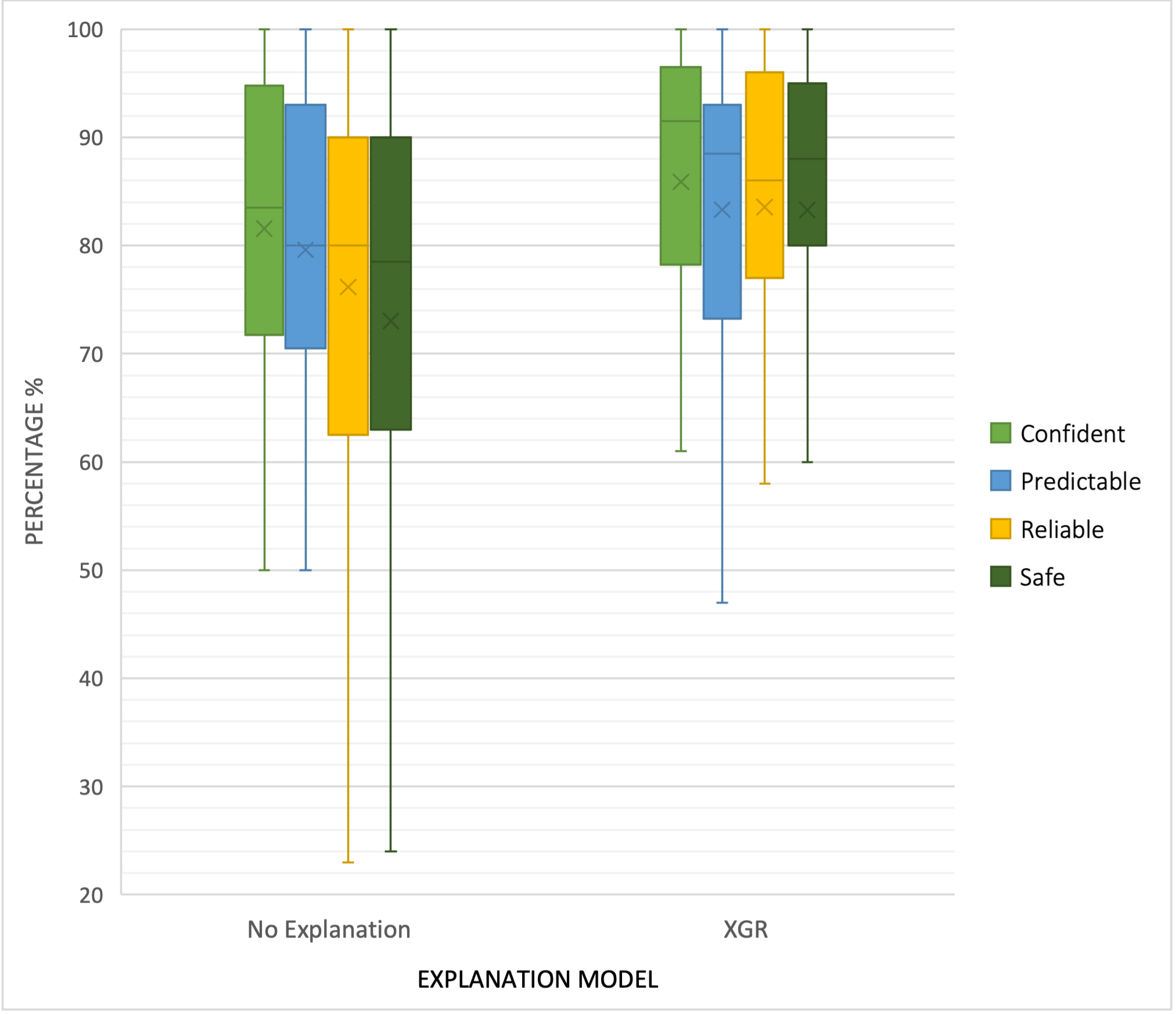} 
\caption{Perceived trust metrics for the two models on the Likert scale percentage with means represented as markers (higher indicates strongly agree).}
\label{fig5}
\end{figure}

Next, we evaluate the second main hypothesis: a better understanding of an agent fosters user trust in the agent. $H_0$ : $T_{XGR} = T_{NoExplanation}$; and $H_1$ : $T_{XGR} \geq T_{NoExplanation}$ where $T$ denotes the participants' perceived trust of the agent. Figure (\ref{fig5}) shows the trust rate variance with the two models. We performed a Welch Two Sample t-test and obtained p-values \textbf{0.12}, \textbf{0.18}, \textbf{0.04}, and \textbf{0.02} for trust metrics \emph{confident}, \emph{predictable}, \emph{reliable}, and \emph{safe} respectively. We reject $H_1$ for the first two metrics and accept it for the rest. Results indicate a significant difference for \emph{reliable}, and \emph{safe} metrics and no difference for \emph{confident}, \emph{predictable}. Although the scores and trust are significantly better for our model, further investigation in a more complex domain is required. 

We also evaluate behavioral trust by measuring the participants' level of agreement with the GR predictions, the difference between the GR prediction, and the participants' correct predictions (task prediction score). Clearly, the trend between the two models shown in Figure \ref{fig4} is the same as for the perceived trust (Figure \ref{fig5}).

\section{Conclusion}
We introduced an explanation model for GR algorithms called eXplainable Goal Recognition XGR. Our model 
generates explanations answering both ‘why’ and ‘why not’ questions pertaining to the problem of GR. Our approach builds upon the WoE concept. 
We computationally evaluated the performance of our system. We also conducted two human studies, showing that our model generates explanations consistent with human labelers in over 73\% of scenarios, and showing that the XGR model improves people's ability to predict an agent's goal, and trust in the GR algorithm. In the future, we plan to extend our work to explain noisy observation sequences, as well as further evaluation of the source of the differences between the model explanation and human-generated explanations. 


\bibliography{aaai22} 


\newpage
 \appendix
 \onecolumn 

\section*{Supplementary Material}
\subsection*{The Weight of Evidence (WoE): Formula Derivation}

The WoE is defined for some evidence $e$, a hypothesis $h$, and its logical complement $\overline{h}$ \cite{wod1985weight} as follows: 

\begin{equation}
woe(h : e) = \log \frac{Odds(h \mid e)}{Odds(h)}  
\end{equation}

\noindent
where the colon is read as ``provided by'', and $Odds(.)$ denotes the hypothesis odds:

\begin{equation}
Odds(h \mid e) = \frac{P(h \mid e)}{P(\overline{h} \mid e)}      \textrm{~~~(Posterior odds)}
\end{equation}

\begin{equation}
Odds(h) = \frac{P(h)}{P(\overline{h})}     \textrm{~~~~~(Prior odds)}
\end{equation}

This is the ratio of the posterior to the prior odds. The odds corresponding to a probability $p$ are defined as $p/(1-p)$ --- the probability of an event occurring divided by the probability of its not occurring.

Using Bayes' rule, $woe(h : e)$ can also be defined as:

\begin{equation}
woe(h : e) = \log \frac{P(e \mid h)}{P(e \mid \overline{h})}
\end{equation}

WoE can also contrast $h$ to an arbitrary alternative hypothesis $h'$ instead of its complement $\overline{h}$. Thus, we can talk generally about the strength of evidence in favor of $h$ and against $h'$ provided by $e$: 

\begin{equation}
woe(h/h' : e) = woe(h : e \mid h \vee h')
\end{equation}

The WoE generalizes to include cases when it can be conditioned on additional information $c$:

\begin{equation}
woe(h : e \mid c) = \log \frac{P(e \mid h, c)}{P(e \mid h', c)}
\end{equation}

From various properties of WoE, the following two properties are essential to our model:

\begin{equation}
\label{eq:woe-h-h-prime}
woe(h/h' : e) =  \log \frac{P(e \mid h)}{P(e \mid h')} 
\end{equation}


\begin{equation}
 \log \frac{P(h)}{P(h')} + \log \frac{P(e \mid h)}{P(e \mid h')} = \log \frac{P(h \mid e)}{P(h' \mid e)}
\end{equation}

By substituting using Equation~\ref{eq:woe-h-h-prime}, we get:

\begin{equation}
 \log \frac{P(h)}{P(h')} + woe(h/h' : e) = \log \frac{P(h \mid e)}{P(h' \mid e)}
\end{equation}

\begin{equation}
woe(h/h' : e) = \log \frac{P(h \mid e)}{P(h' \mid e)} - \log \frac{P(h)}{P(h')} 
\end{equation}

Using the log quotient property, we simplify the equation as follows: 

\begin{equation}
woe(h/h' : e) = \log \frac{\frac{P(h \mid e)}{P(h' \mid e)}}  {\frac{P(h)}{P(h')} }
\end{equation}

If we have uniform prior probabilities, we can simplify further and compute WoE for a pair of hypotheses (conditioned on $c$) as follows:

\begin{equation}
woe(h/h' : e \mid c) = \log \frac{P(h \mid e, c)}{P(h' \mid e, c)}  
\end{equation}

\bigskip
\end{document}